\documentclass[11pt]{article}

\usepackage{jmlr2e}
\usepackage[nosumlimits,nonamelimits]{amsmath}
\usepackage{cleveref}
\usepackage[utf8]{inputenc} %
\usepackage[T1]{fontenc}    %
\usepackage{hyperref}       %
\usepackage{url}            %
\usepackage{booktabs}       %
\usepackage{amsfonts}       %
\usepackage{nicefrac}       %
\usepackage{microtype}      %
\usepackage{xcolor}         %
\usepackage{adjustbox}
\usepackage{enumitem}
\usepackage{multirow}

\usepackage{subfigure}

\usepackage{caption}

\usepackage{tikz}
\usetikzlibrary{fit,positioning}

\usepackage{dsfont}

\usepackage{graphicx} %
\usepackage{amssymb}
\usepackage{times}
\usepackage{epsfig}
\usepackage{graphicx}
\usepackage{color}
\usepackage{bbm}
\usepackage{multicol}
\usepackage{natbib}
\usepackage{wrapfig}

\providecommand{\hide}[1]{}

\usepackage{rotating}

\usepackage{amsthm}
\usepackage{amsopn,amssymb,thmtools,thm-restate}

\usepackage{bm} %
\usepackage{algorithm}%
\usepackage{algpseudocode}%
\newcommand{\vct}[1]{\boldsymbol{#1}} %
\newcommand{\mat}[1]{\boldsymbol{#1}} %

\newcommand{\field}[1]{\mathbb{#1}}
\newcommand{\R}{\field{R}} %
\newcommand{\Z}{\field{Z}} %
\newcommand{\T}{^{\top}} %

\newcommand{\ProbOpr}[1]{\mathbb{#1}}

\newcommand{\expect}[2]{%
\ifthenelse{\equal{#2}{}}{\ProbOpr{E}_{#1}}
{\ifthenelse{\equal{#1}{}}{\ProbOpr{E}\left[#2\right]}{\ProbOpr{E}_{#1}\left[#2\right]}}} %
\newcommand{\var}[2]{%
\ifthenelse{\equal{#2}{}}{\ProbOpr{VAR}_{#1}}
{\ifthenelse{\equal{#1}{}}{\ProbOpr{VAR}\left[#2\right]}{\ProbOpr{VAR}_{#1}\left[#2\right]}}} %

\DeclareMathOperator{\argmax}{arg\,max}

\newcommand{\vmu}{\vct{\mu}}

\newcommand{\vx}{{\vct{x}}}
\newcommand{\vy}{\vct{y}}

\newcommand{\mI}{\mat{I}}

\def\N{{\mathcal{N}}}
\def\GP{\mathcal{GP}}

\def\KL{\textrm{KL}}

\def\tr{\textrm{tr}}

\usepackage{xspace}

\newcount\Comments  %
\usepackage{color}
\definecolor{darkred}{rgb}{0.5,0,0}
\definecolor{orange}{rgb}{1.0,0.64,0}
\definecolor{darkgreen}{rgb}{0,0.5,0}
\definecolor{darkblue}{rgb}{0,0,0.7}
\definecolor{purple}{rgb}{.6, 0,.6}
\newcommand{\kibitz}[2]{\ifnum\Comments=1\textcolor{#1}{#2}\fi}

\definecolor{darkblue}{rgb}{0.0,0.0,0.55}
\hypersetup{
  colorlinks = true,
  citecolor  = darkblue,
  linkcolor  = darkblue,
  citecolor  = darkblue,
  filecolor  = darkblue,
  urlcolor   = darkblue,
}

\newcommand{\hyperbo}{HyperBO\xspace}

\jmlrheading{}{All authors are affiliated with Google Research, specifically the Brain Team.}{}{}{}{}

\firstpageno{1}

\begin{document}
\title{Pre-training helps Bayesian optimization too}
\author{\name Zi Wang \email wangzi@google.com 
       \AND
       \name George E. Dahl \email gdahl@google.com 
       \AND 
       \name Kevin Swersky \email kswersky@google.com
       \AND 
       \name Chansoo Lee \email chansoo@google.com
       \AND 
       \name Zelda Mariet \email zmariet@google.com
       \AND 
       \name Zachary Nado \email znado@google.com
       \AND 
       \name Justin Gilmer \email gilmer@google.com
       \AND 
       \name Jasper Snoek \email jsnoek@google.com
       \AND 
       \name Zoubin Ghahramani \email zoubin@google.com
       }
\editor{}

\maketitle
\vspace{-2em}
\begin{abstract}
Bayesian optimization (BO) has become a popular strategy for global optimization of many expensive real-world functions. Contrary to a common belief that BO is suited to optimizing black-box functions, it actually requires domain knowledge on characteristics of those functions to deploy BO successfully. Such domain knowledge often manifests in Gaussian process priors that specify initial beliefs on functions. However, even with expert knowledge, it is not an easy task to select a prior. This is especially true for hyperparameter tuning problems on complex machine learning models, where landscapes of tuning objectives are often difficult to comprehend. We seek an alternative practice for setting these functional priors. In particular, we consider the scenario where we have data from similar functions that allow us to pre-train a tighter distribution a priori. To verify our approach in realistic model training setups, we collected a large multi-task hyperparameter tuning dataset by training tens of thousands of configurations of near-state-of-the-art models on popular image and text datasets, as well as a protein sequence dataset. Our results show that on average, our method is able to locate good hyperparameters at least 3 times more efficiently than the best competing methods.

\end{abstract}

\section{Introduction}
Bayesian optimization (BO) has been successfully applied in numerous real-world global optimization problems, ranging broadly from hyperparameter tuning~\citep{snoek2012practical, kotthoff2019auto} to chemical synthesis~\citep{shields2021bayesian, griffiths2020constrained}, drug discovery~\citep{pyzer2018bayesian}, aerospace engineering~\citep{lam2018advances}, robotics~\citep{driess2017constrained,wang17icra} and the list goes on. 
These successful BO applications benefit from expert knowledge on characteristics of the function to be optimized and hands-on experience with BO on similar tasks in the past. Such knowledge or experience can give intuitions about a functional form of the problem and thus specifications of a functional prior~\citep{turner2021bayesian}. However, without domain knowledge or hands-on experience, BO performance can be susceptible to misspecified priors~\citep{schulz2016quantifying} unless the misspecification is insignificant~\citep{bogunovic2021misspecified}. 

Similar to how practitioners learn how to set good priors from past experience, we automate the prior determination process by \emph{pre-training priors} on data that are available on different but related tasks. 
Given the benefit of pre-trained priors on synthetic functions, simple tuning tasks and complex robotics tasks~\citep{wang2018regret, kim2019learning, perrone2018scalable}, can we take it to the level of real-world hyperparameter tuning problems for modern deep learning models (e.g. ResNet50) and large-scale datasets (e.g. ImageNet)? To the best of our knowledge, there is no such multi-task tuning benchmark available for modern large models and datasets, but this kind of tuning task is most prevalent in hyperparameter tuning in recent years of ML-related publications given the success of large models~\citep{he2016deep, raffel2019exploring, brown2020language}. For these problems, it is difficult to understand the landscapes of tuning objectives, hindering the use of Bayesian optimization with almost unobtainable expert interventions on priors. 
To fill the vacancy of a dataset for hyperparameter tuning in modern ML, we collected a large multi-task hyperparameter tuning dataset by training tens of thousands of configurations of near-state-of-the-art models on popular image and text datasets, as well as on a protein sequence dataset. Our open-sourced dataset can save roughly 12,000 machine-days of computation for anyone who makes use of it.

On the modeling side, most existing meta BO methods either scale cubically in the number of evaluations and tasks~\citep{swersky2013multi, bardenet2013collaborative}, impose a restrictive set of assumptions on the available data~\citep{wang2018regret, swersky2013multi} for efficient solutions, or make assumptions on the availability of GP parameters~\citep{volpp2020meta} or descriptive task-level features~\citep{brazdil1994characterizing, bardenet2013collaborative, yogatama2014efficient}. 
To address these issues, we introduce \hyperbo: a meta BO method that builds upon~\citet{wang2018regret} with a simple assumption: all the related functions being optimized are samples from the same GP prior distribution.  Concretely, \hyperbo assumes the functions are conditionally independent given the hyperparameters, mean and covariance function of the GP.  %
Compared to~\citet{wang2018regret}, \hyperbo does not impose any strict conditions on data or model structures. From a computational perspective, \hyperbo scales linearly in the number of tasks during training, and does not depend on the number of tasks when deployed.  By not imposing assumptions about the data collection conditions, it can be used with large offline datasets or a few related optimization trajectories.

Our empirical results show that \hyperbo is at least 3 times more efficient in function evaluations than recent baseline methods to locate the best hyperparameters. Our main contributions are two-fold: (1) a new method for BO with learned (as opposed to hand-designed) priors; (2) a large multi-task hyperparameter tuning dataset that not only benefits our method but also serves as a realistic benchmark to test future methods. Both open-sourced code and dataset are available at \url{https://github.com/google-research/hyperbo}. Our extended paper~\citep{wang2022pre} is available for interested readers to deep dive.

\section{Problem formulation}
\label{sec:pf}
We consider the standard black-box function optimization scenario: given a real-valued function $f$ defined over a compact, hyper-rectangular space $\mathfrak X \subset \R^d$ and given observations of similar functions $f_1, \cdots, f_N$, we seek an $x \in \mathfrak X$ optimizing $f$. We inherit our problem formulation from~\citet{wang2018regret}, but we relax impractical assumptions on data availability (we do not require all observations to be made on the same inputs across tasks) and model restrictions.%

\paragraph{Assumptions and the goal.}Concretely, we assume that there exists a Gaussian process $\GP(\mu, k)$ with unknown mean function~$\mu: \mathfrak X\rightarrow \R$ and kernel~$k: \mathfrak X\times \mathfrak X\rightarrow \R$. Let $N$ be the number of tasks and let $M_i$ be the number of observations we have for the $i$th task. Conditioned on independent function samples $f_i\sim \GP(\mu, k)$ and inputs $x^{(i)}_{j}\in \mathfrak X, i\in[N], j\in[M_i]$, we observe evaluations $y^{(i)}_{j}\sim \N(f_i(x^{(i)}_{j}), \sigma^2)$ perturbed by \textit{i.i.d.\ }additive Gaussian noise $\N(0, \sigma^2)$  with unknown variance $\sigma^2$. Taken together, the collection of sub-datasets $D_{f_{i}} = \{(x^{(i)}_{j}, y^{(i)}_{j})\}_{j=1}^{M_i}$ define a dataset $D_N = \{D_{f_{i}} \}_{i=1}^N$. Finally, our goal is to maximize a new function independently sampled from the same GP, $f\sim \GP(\mu, k)$; that is, solve $\underset{x\in \mathfrak X}{\argmax} \; f(x)$ given dataset $D_N$ but unknown functions $\mu,k$ and unknown parameter $\sigma^2$.

\paragraph{An example.} In our optimizer hyperparameter tuning application, a task corresponds to finding the best optimizer hyperparameters to train a given model on a particular dataset,\footnote{Technically, we also consider different batch sizes to be different tasks.} e.g. training a ResNet~\citep{he2016deep} on ImageNet~\citep{imagenet}. Notice that we do not assume that the mean function $\mu$, kernel $k$ and noise variance $\sigma^2$ are given. This is consistent with the reality of solving real-world black-box optimization problems including hyperparameter tuning. We must learn those unknown functions and parameters from data. However, in practice, searching in functional spaces to find the right mean $\mu$ or kernel $k$ is a daunting task. Hence for practical concerns, a well defined search space for functions is required. More details on this can be found at \S\ref{ssec:nll}.

\paragraph{Metrics.}
For simplicity, throughout this paper, we focus on the setting where the target function $f$ can only be optimized by iteratively choosing where to evaluate, and defer batch evaluation setups to Sec.~\ref{sec:discuss}. %
As we run BO on the target function $f$ for $T$ iterations, we accumulate a set of observations $D_f = \{(x_t, y_t)\}_{t=1}^T$, $y_t \sim \mathcal N(f(x_t), \sigma^2)$. We evaluate the quality of the optimization using the \emph{simple regret} metric: $R_T = \max_{x\in \mathfrak X} f(x) - f(\hat x)$, where $\hat x$ is the final recommendation at the end of the optimization process. There are various ways of setting $\hat x$ based on the observations $D_f$; we use the input that achieved the best evaluation: $\hat x = x_\tau; \tau = \argmax_{t\in[T]}y_t$. %

\paragraph{Notations.} Let $[n]$ denote $\{1,\cdots,n\}, \forall n\in \Z^+$. For conciseness, we write the evaluation of a function $f: \mathbb R \to \mathbb R$ on vector $\vx = [x_i]_{i=1}^{n}$  as $\mu(\vx) := [\mu(x_i)]_{i=1}^n$. Similarly, for two vectors $\vx, \vx'$, we write the corresponding kernel matrix as $k(\vx, \vx') := [k(x_i, x'_j)]_{i\in[n], j\in[n']}$, and shorten $k(\vx) := k(\vx,\vx)$. 

We denote a (multivariate) Gaussian distribution with mean $u$ and variance $\Sigma$ by $\N(u, \Sigma)$, and a Gaussian process (GP) with mean function $\mu$ and covariance function $k$ by $\GP(\mu, k)$. Let $\sigma^2$ be the noise variance in observations. Given a set of observations $D=\{(x_t, y_t)\}_{t=1}^T, \vy_T = [y_t]_{t=1}^T\sim \N(f(\vx_T),\sigma^2 \mI), \vx_T=[x_t]_{t=1}^T$ and $f\sim \GP(\mu, k)$, 
we denote the corresponding conditional GP distribution as $\GP(\mu, k\mid D)$. Recall that the conditional distribution $\GP(\mu, k\mid D)=\GP(\mu_D, k_D)$, is given for any $x, x' \in \mathfrak X$ as
\begin{align} \label{eq:conditionalgp}
\mu_D(x) &=\mu(x) + \psi(x)(\vy_T - \mu(\vx_T)), \\\
k_D(x,x') &= k(x,x') - \psi(x) k(\vx_T, x'),
\end{align}
where we set  $\psi(x)= k(x, \vx_T)(k(\vx_T)+\sigma^2\mI)^{-1}$. %

\section{Our method}
\label{sec:method}
\begin{wrapfigure}{R}{0.55\textwidth}
    \begin{minipage}{0.55\textwidth}
 \begin{algorithm}[H]
       \caption{HyperBO with acquisition function~$\alpha(\cdot)$.}\label{alg:hyperbo}
  \begin{algorithmic}[1]
    \Function{HyperBO\,}{$f, D_N$}
    \State $\GP(\hat \mu, \hat k)\gets \textsc{Pre-Train}(D_{N})$\label{alg:train}
    \State $D_f \gets \emptyset$
    \For{$t = 1,\cdots, T $} \label{alg:bostart}
        \State $x_t\gets \underset{x\in\mathfrak X}{\argmax}{\,\alpha\left(x ; \GP(\hat \mu, \hat k \mid D_f ) \right)}$ \label{alg:strategy}
        \State $y_t\gets$ \textsc{Observe}$\left(f(x_t)\right)$
        \State $ D_f \gets D_{f}\cup \{(x_t,y_t)\}$
      \EndFor \label{alg:boend}
     \State \Return $D_f$
    \EndFunction
  \end{algorithmic}
\end{algorithm}
    \end{minipage}
  \end{wrapfigure}
As shown in~Alg.~\ref{alg:hyperbo}, our approach pre-trains the GP hyperparameters on a representative set of datasets and fixes them for the duration of the optimization procedure; we refer to this approach as HyperBO. HyperBO runs in two steps. First, we learn a GP model $\GP(\hat \mu, \hat k)$ to approximate the ground-truth (unknown) GP that generated the dataset $D_N$. Then, we do standard BO to optimize a new function $f$ with the learned GP $\GP(\hat \mu, \hat k)$. The initial pre-training process (Alg.~\ref{alg:hyperbo}, line~\ref{alg:train}) is the critical difference between HyperBO and standard BO algorithms, as well as the key contribution of this paper.  %

In each iteration of the BO loop (Alg.~\ref{alg:hyperbo}, lines~\ref{alg:bostart}-\ref{alg:boend}), we update the conditional GP, but do not re-estimate the GP mean and kernel. 
By separating the data for conditional GP update and GP parameter pre-training, we minimize the computational cost while still maintaining good performance empirically. Moreover, we avoid the BO chicken-and-egg dilemma~\citep{wang2018regret} where the search strategy is trained on data collected in the BO process and the data points are selected by the search strategy simultaneously.

Next, we introduce our GP pre-training strategy based on two types of objectives: KL divergence between estimates and model predictions (\S~\ref{ssec:reg}) and negative log likelihood (\S~\ref{ssec:nll}). 

\subsection{Pre-training with empirical KL divergence}
\label{ssec:reg}
We first investigate the case where observations on the same set of inputs are available. This is the main scenario considered by \citet{wang2018regret}, but their method only works for Bayesian linear regression. We now present \hyperbo based on an empirical KL divergence to allow highest flexibility on mean functions and kernels, e.g. a Mat\'ern kernel on deep features shared with a mean function. The objective is called \emph{empirical KL divergence} because it is the KL divergence between an empirically estimated multivariate Gaussian and model predictions from a GP.

Here we consider a special case of dataset $D_N$ which contains matching inputs across some tasks. More formally, suppose we have a \emph{matching dataset} $D'_N = \{(x_j, \vy_j)\}_{j=1}^M$ where $M$ is the number of shared inputs across N tasks. For each input index $j\in [M]$ and input $x_j\in\mathfrak X,$ we have $N$ observed values $\vy_j = [y_{j}^{(i)}]_{i=1}^{N} \in \R^{N}$ and each observation $y_{j}^{(i)} \sim \N(f(x_j), \sigma^2)$ corresponds to evaluating input $x_j$ on a different task. In practice, dataset $D_{N}'$ can be constructed by querying a set of functions $f_{1}, \cdots, f_{N}$ at the same set of input locations $\vx = [x_j]_{j=1}^{M} \in \R^{M\times d}$ to obtain an observation matrix $\vy =[\vy_j]_{j=1}^M \in \R^{M\times N}$.

By definition of a GP, the vector of all function queries $f(\vx)$ is distributed according to a multivariate Gaussian distribution $\N(\mu(\vx), k(\vx))$. With our observation model, we get the distribution for observations $\vy \sim \N(\mu(\vx), k(\vx) +\mI\sigma^2)$ for some unknown mean function $\mu$ and kernel $k$. 

However, given that we have access to all observations $\vy$, we can estimate the mean on inputs $\vx$ as $\tilde \vmu = \frac{1}{N} \vy 1_{N} \in \R^M$ and estimated covariance as $\tilde K = \frac{1}{N}(\vy -\tilde \vmu1_N\T) (\vy- \tilde \vmu1_N\T)\T \in\R^{M\times M}$; here $1_N$ is a column vector of size $N$ filled with $1$s. We use a biased estimate of covariance to be consistent with the maximum likelihood estimator in \S\ref{ssec:nll}.  But one may choose to re-scale learned kernel by $\frac{N}{N-1}$ to be unbiased. Notice that the estimated covariance includes in diagonal terms the variance of the observation noise. 

For any divergence function between the estimate $\N(\tilde \vmu, \tilde K)$ and model prediction $\N(\mu(\vx), k(\vx) +\mI\sigma^2)$, we obtain an objective to minimize, $\mathcal D\left(\N(\tilde \vmu, \tilde K), \N(\mu(\vx), k(\vx) +\mI\sigma^2)\right)$. While there are different measures of distributional discrepancy, we adopt the KL divergence. Let $\vmu = \mu(\vx)$ and $K = k(\vx) +\mI\sigma^2$. The empirical KL divergence is defined as 
{\small
\begin{align}\label{eq:kl}
    &\mathcal D_{\KL}\left(\N(\tilde \vmu, \tilde K), \N(\vmu, K)\right) =\frac12\left(\tr(K^{-1}\tilde K) + (\vmu - \tilde \vmu)\T K^{-1}(\vmu - \tilde \vmu) + \ln\frac{|K|}{|\tilde K|} - M \right),
\end{align}
}%
and we can estimate the mean, kernel and noise variance by minimizing $\mathcal D_{\KL}$.

\subsection{Pre-training with negative log likelihood}
\label{ssec:nll}
If we have arbitrary data points from each task, a straightforward way to pre-train a GP is by optimizing a negative log likelihood (NLL) over parameters of the GP. The regression-based NLL objective corresponds to how supervised pre-training is done with a cross-entropy loss on deep learning models for classification tasks. Here, we use the NLL on the given observations from multiple functions that are assumed to be independently sampled from the GP. The independence naturally results in a summation over NLLs for all observed functions, which is a key difference to the widely used type II maximum likelihood approximation for GP inference on a single function in BO setups. The NLL loss function for our method is
$L(\mu, k, \sigma^2) = -\sum_{i=1}^N\log p(D_{f_i} \mid \mu, k, \sigma^2)$. 
We then obtain a solution to the choice of mean function, kernel function and noise variance by minimizing the NLL loss function.

\section{Experiments}
\label{sec:exp}

Our goal in this paper is to provide a practical approach for hyperparameter optimization when we are given data on a range of tasks over the same search space. To analyze the effectiveness of our proposal, we take the optimizer hyperparameter tuning problem in deep learning as a case study. We collected a dataset composed of hyperparameter evaluations on various deep neural network training tasks. The tasks included optimizing deep models on image, text, and other datasets. Our code and dataset can be found at \url{https://github.com/google-research/hyperbo}. 

To reduce ambiguity, we distinguish between datasets that individual neural networks are trained on and the dataset we collected that includes optimizer hyperparameter points with their validation errors (and other metrics). We will call the former (e.g. MNIST, CIFAR10) task datasets and call the latter the tuning dataset. The tuning dataset is what we described as dataset $D_N$ in \S\ref{sec:pf}.

For \hyperbo, we used thresholded probability of improvement as the acquisition function and GP with a one-hidden layer neural network of size 8 as mean function and a Mat\'ern32 covariance on the feature layer of the mean function as kernel. H* NLL refers to HyperBO with GPs pre-trained via the NLL objective in \S\ref{ssec:nll} and H* KL refers to pre-training via the KL objective in \S\ref{ssec:reg}.

Our baselines include (1)~Rand: Random search in the corresponding scaled search space. (2)~STBO: Single-task BO where in every BO iteration, STBO uses the same GP architecture as \hyperbo but optimizes the GP hyperparameters via type II maximum likelihood on data of the test task. This implementation corresponds to the basic off-the-shelf BO setups. (3)~STBOH: Single-task GP-UCB with a \emph{hand-tuned} prior on hyper-parameters including UCB coefficient~\cite{srinivas2009gaussian, Golovin2017}. (4)~MIMO: Multi-task BO with GP bases as an ensemble of feedforward neural networks with shared subnetworks~\citep{kim2021scalable,havasi2020training}. (5)~RFGP: Multi-task BO with GP bases as random features~\citep{snoek2015scalable, krause2011contextual}. (6)~MAF: Meta acquisition function method based on reinforcement learning~\cite{volpp2020meta}, which assumes the knowledge of the best GP hyperparameters for each task.

\begin{figure*}
    \centering
    \includegraphics[width=1.\textwidth]{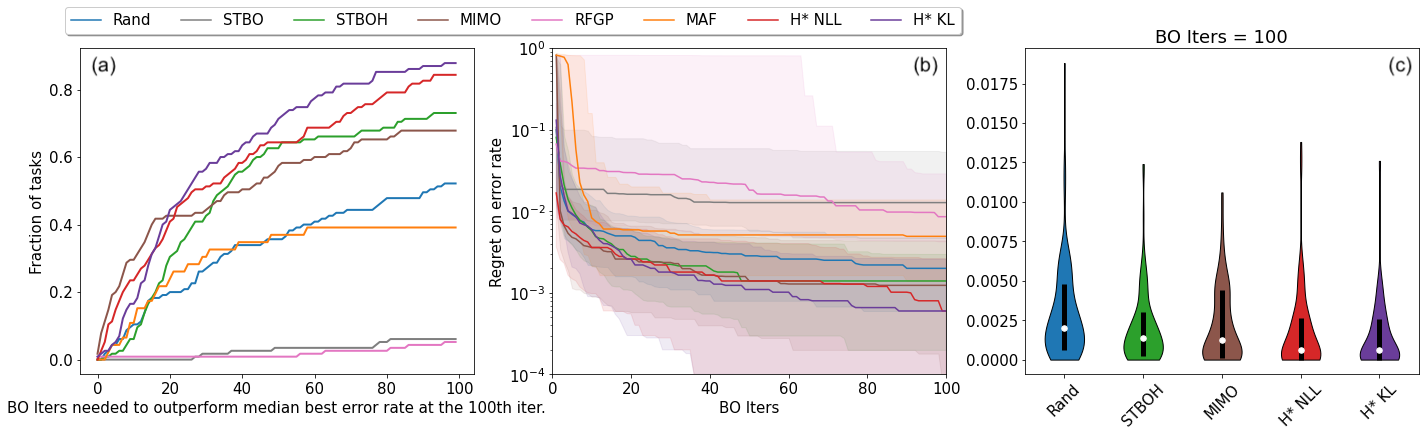}
    \caption{(a) The performance profile for outperforming the median of best error rates at the 100th BO iteration. (b) BO convergence of all methods: the median and 20/80 percentiles of the regrets on error rates over 115 BO runs: 23 tasks and each with 5 repeats of different random seeds. (c) A violin plot on the vertical slices of (b) at the 100th iteration; the white dot is the median and the black line is the 20/80 percentile. Overall, \hyperbo methods H* NLL and H* KL are able to achieve the lowest regret on error rate on the majority of tasks with fewer iterations.}
    \label{fig:hold-out-related-summary-iclr}
\end{figure*}

Fig.~\ref{fig:hold-out-related-summary-iclr}~(a) shows the \emph{performance profiles}, the fraction of all test tasks that each method is able to outperform a baseline criterion at each BO iteration. MIMO is able to outperform other methods in the beginning 20 BO iterations, but its leading position soon gets surpassed by \hyperbo (H* NLL and H* KL). Fig.~\ref{fig:hold-out-related-summary-iclr}~(b,c) illustrates the BO convergence curves of all competing methods, together with the vertical slice at the 100th iterations. MAF, RFGP and STBO are both falling much behind Rand. Surprisingly, the contextual information learned by RFGP did not generalize to a new task. On the other hand, MIMO is able to obtain a slightly better error rate than STBOH. Overall, learning the GP prior through data as with \hyperbo outperforms other meta BO methods, and is a more principled and effective way to obtain the GP prior when compared with hand-tuning.

To more precisely quantify \hyperbo's advantage, we also computed how much faster \hyperbo can get a better error rate than best alternatives, which can be different from task to task. We found that on average, on over $50\%$ tasks, H* NLL is at least 2.86 times faster than best non-\hyperbo alternatives; while on over $57\%$ tasks, H* KL is at least 3.26 times faster than best non-\hyperbo alternatives. Moreover, on over $73\%$ tasks, H* NLL is at least 7.74 times faster than random search; and on over $75\%$ tasks, H* KL is at least 6.07 times faster than random search.

\section{Conclusion}
\label{sec:conclu}
We proposed \hyperbo: a novel meta BO approach that supports practical applications that involve continuous inputs queried at possibly non-aligned locations across tasks. \hyperbo uses a simple yet effective idea that is easy to implement and efficient to run. We evaluated \hyperbo on real-world big model optimizer tuning tasks, and the results demonstrated its superior performance over state-of-the-art competing methods.

\bibliography{refs}

\appendix
\section{Related work}

There is a rich literature of innovative methodologies to improve the efficiency of BO given related tasks or additional context.  Here we discuss the most closely related work and explain why these don't solve the specific scenario which we envision.  Specifically, our goal is a methodology that is scalable enough to share information across thousands of tasks, each with potentially hundreds of observations, such as in the context of a large BO service or library.

Pre-training and prior learning is directly related to meta learning, learning to learn and learning multiple tasks~\citep{Baxter1996}. We use the word pre-training to refer to supervised pre-training, which is a general approach in the deep learning community~\citep{girshick2014rich} to transfer knowledge from prior tasks to a target task. The same as pre-training deep features on a variety of tasks, \citet{wang2018regret} proposed prior learning for GPs to learn the basis functions by treating the independent function outputs as individual heads of a neural network.

Several methods, including that which \hyperbo extends, refer to their method as ``meta-BO''~\citep{wang2018regret, volpp2020meta}. However, in this work we use the term \emph{meta-BO} more generally to refer to the class of BO methods that use data from existing tasks to optimize a new task.  Since standard BO is a learning process, it is consistent to call those methods meta BO methods given that they learn how to learn. Under this viewpoint, meta BO approaches also include multi-task BO~\citep{swersky2013multi, poloczek2017multi, yogatama2014efficient}, transfer learning BO using contextual GPs~\citep{krause2011contextual, bardenet2013collaborative, poloczek2016warm} and transfer learning based on quantiles~\citep{salinas2020quantile}. Some meta BO methods have also been studied for hyperparamter tuning tasks in machine learning~\citep{feurer2015efficient, salinas2020quantile}.

\hyperbo assumes all tasks are independent (after conditioning on the GP), whereas both multi-task and contextual BO rely heavily on the assumption that tasks are related.  Thus the latter approaches typically scale cubically in both the number of tasks and observations in each task, meaning that they cannot gracefully scale across both without heavy approximations. When assuming that all inputs are equal across tasks, multi-task BO can be sped up using a Kronecker decomposition of the kernel to a task kernel and an input kernel which can be inverted separately; a similar assumption is made by~\citet{wang2018regret}.  In comparison, \hyperbo scales linearly in the number of tasks.

Another thread of meta BO literature was started in the robot learning area by \citet{kim2017learning,kim2019learning}, which estimated a multivariate Gaussian to transfer knowledge on scoring functions for search strategies in robot manipulation tasks, and thus only considered finite discrete inputs. \citet{wang2018regret} provided regret bounds for \citet{kim2017learning,kim2019learning} and extended it to continuous search spaces by considering a GP as a Bayesian linear regressor with neural net basis functions.

Similar ideas were adopted by \citet{perrone2018scalable, wistuba2021few} in the machine learning hyperparameter tuning literature. These ideas, on a high level, can be viewed as special cases of \hyperbo or~\citet{wang2018regret}. Compared to \citet{wang2018regret}, \citet{perrone2018scalable} and \citet{wistuba2021few} only use zero means, which, as shown by \citet{kim2017learning,kim2019learning}, is critical for learning the initial data points to acquire. As a remedy, \citet{wistuba2021few} developed an evolutionary algorithm based data-driven strategy to warm start the initialization of data selection. Although different terms are used, \citet{wang2018regret} and \citet{perrone2018scalable} concurrently proposed the idea of learning parameters of GP priors from multi-task datasets, while \citet{wang2018regret} is the first to clarify the assumptions that those multi-task functions need to be conditionally independent so that regret bounds hold for BO with an unknown GP prior.

\hyperbo builds upon~\citet{wang2018regret} and~\citet{kim2017learning,kim2019learning}, yet principally resolves their limitations on search spaces and data availability. For both finite discrete search spaces and continuous ones, \citet{wang2018regret} requires observations on the same set of inputs across tasks, which is an assumption that is not required for \hyperbo. Both arbitrary data points from different tasks and observations on same inputs across tasks can be incorporated into \hyperbo efficiently and effectively. Another critical advantage of \hyperbo is accommodations of very flexible kernels and mean functions that are not limited by Bayesian linear regressors; this opens meta BO to a lot more GP architectures involving combinations of deep features and kernels with infinite basis functions.

\section{Discussion}
\label{sec:discuss}
In this work, we focused on the question of how to efficiently and effectively make use of multi-task data to enable better BO with pre-trained priors. We simplified other aspects of BO that are orthogonal to our focuses, such as parallel queries or different search spaces. Here we discuss extensions to our work that would enable even more flexible uses.

\paragraph{Batch evaluation.}
For simplicity of this paper, we did not consider batch evaluation but rather only focused on the prior selection dimension of the challenges in BO. However, it is straightforward to adopt any batch BO methods in conjunction with \hyperbo to support obtaining observations in parallel. For example, we can directly use batch methods from \citet{snoek2012practical, kathuria2016batched, wang2017batched} etc. to replace line~\ref{alg:strategy} of Alg.~\ref{alg:hyperbo}.

\paragraph{High-dimensional and large scale data.}
Similar to batch BO, our method can also be naturally combined with most high-dimensional and large scale BO methods to offer more capabilities. For these cases, typically a probabilistic model different from vanilla GPs may be adopted. In line~\ref{alg:train} of Alg.~\ref{alg:hyperbo}, it is straightforward to adapt our method to optimize the cumulative marginal likelihood in \S\ref{ssec:nll} instead for the new model. Our meta-learning idea in this paper in fact also brings benefit to high-dimensional and large scale BO methods so that they can better identify their critical special structures, e.g. low-dimensional embedding~\citep{wang2016bayesian}, cylindrical kernels~\citep{oh2018bock} or additive Mondrian kernels~\citep{wang2018batched}.

\paragraph{Different search spaces.}
Roughly speaking, there could be two circumstances for difference search spaces. Case I is that tasks share the same search variables, but the search ranges for some variables are different. For example, we may have each function $f_i:\mathfrak X_i \rightarrow \R, i \in [N]$ and $\mathfrak X_i = \prod_{j=1}^d[l_{ij}, h_{ij}] \subset \R^d$. In this case, our solution still applies by simply setting a union search space as $\mathfrak X = \bigcup_{i=1}^N \mathfrak X_i$ for learning and use the designated search space of new tasks for optimization.

Case II is more complicated: the search space for each function $f_i$ is $\mathfrak X_i \subset \R^{d_i}$ and each dimension of $\mathfrak X_i$ may have a different meaning than another search space $\mathfrak X_j$ ($i\neq j$). This paper does not have a solution for this scenario. Further research will be needed to reduce Case II to Case I which can be then immediately combined with \hyperbo.

\end{document}